# Joint Correction of Attenuation and Scatter Using Deep Convolutional Neural Networks (DCNN) for Time-of-Flight PET

Jaewon Yang[1], Dookun Park[2], Jae Ho Sohn[1], Zhen Jane Wang[1], Grant T. Gullberg[1], Youngho Seo[1]

[1]Physics Research Laboratory, Department of Radiology and Biomedical Imaging, University of California, San Francisco, CA
[2]Microsoft, Bellevue, WA

*Abstract*–Deep convolutional neural networks (DCNN) have demonstrated its capability to convert MR image to pseudo CT for PET attenuation correction in PET/MRI. Conventionally, attenuated events are corrected in *sinogram* space using attenuation maps derived from CT or MR-derived pseudo CT. Separately, scattered events are iteratively estimated by a 3D model-based simulation using down-sampled attenuation and emission sinograms. However, no studies have investigated joint correction of attenuation and scatter using DCNN in *image* space. Therefore, we aim to develop and optimize a DCNN model for attenuation and scatter correction (ASC) simultaneously in PET *image* space without additional anatomical imaging or time-consuming iterative scatter simulation. For the first time, we demonstrated the feasibility of directly producing PET images corrected for attenuation and scatter using DCNN ($PET_{DCNN}$) from noncorrected PET ($PET_{NC}$) images.

## I. Introduction

Attenuation and scatter correction (ASC) is critical for quantitative accuracy as well as image quality in PET [1]. Attenuated and scattered events occur due to photoelectric effects and Compton scattering induced by the presence of dense material along lines of response (LORs). Without attenuation correction, regions near the skin appear darker (emitting more photons) and regions surrounding brain tissues appear brighter (emitting less photons). Scatter fraction can reach 50% to 60% of LORs recorded in whole-body 3D PET and, without scatter correction, LORs recorded outside an object boundary due to scatter contribute noise in image reconstruction. Therefore, it is important to compensate for attenuation and scatter for quantitative PET.

In a hybrid PET/CT or PET/MRI, attenuation maps are generated from CT [2] or MR-derived pseudo CT images [3, 4] for attenuation correction; while, scattered events are iteratively estimated for scatter correction by a 3D model-based simulation using down-sampled attenuation and emission sinograms [5]. Both attenuation and scatter correction are separately performed due to the difference of photoelectric effects and Compton scattering in *sinogram* space where LORs are conventionally recorded as counts according to their locations and orientations [1].

Recently, deep convolutional neural networks (DCNN) are being actively investigated for PET applications such as attenuation map generation [6-9], image denoising [10, 11] and reconstruction [12]. However, no studies have investigated DCNN-based attenuation or scatter correction using only PET images. Therefore, we aim to develop and optimize a DCNN model for joint ASC in PET *image* space (Figure 1) without additional anatomical imaging or time-consuming iterative scatter simulation.

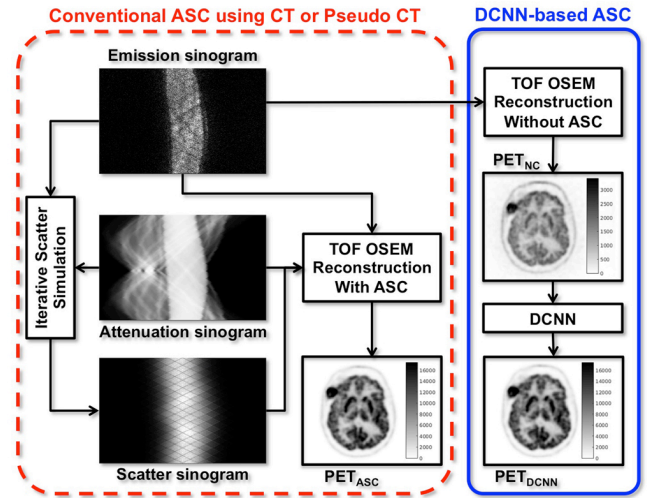

**Figure 1**. Schematic of conventional attenuation and scatter correction (ASC) performed in *sinogram* space during PET image reconstruction (left) and proposed deep convolutional neural network (DCNN)-based ASC performed in *image* space (NC: noncorrected).

## II. Method and Materials

### A. Data Set

$^{18}$F-FDG brain PET images were used for developing and optimizing a DCNN architecture for PET neuroimaging. 35 subjects (25/10 split for training and test data) underwent helical CT (Discovery PET/CT, GE Healthcare; Biograph HiRez 16, Siemens Healthcare) followed by Time-of-flight (TOF)-PET/MRI (SIGNA, GE Healthcare) for 227.2 ± 137.5 s

This work was supported in part by the National Institutes of Health under Grants R01HL135490 and R01EB026331.

J. Yang is with Physics Research Laboratory, Department of Radiology and Biomedical Imaging, University of California, San Francisco, CA 94143 USA (telephone: 415-353-4910, e-mail: Jaewon.yang@ucsf.edu).

Y. Seo is with Physics Research Laboratory, Department of Radiology and Biomedical Imaging, University of California, San Francisco, CA 94143 USA (telephone: 415-353-9464, e-mail: youngho.seo@ucsf.edu).

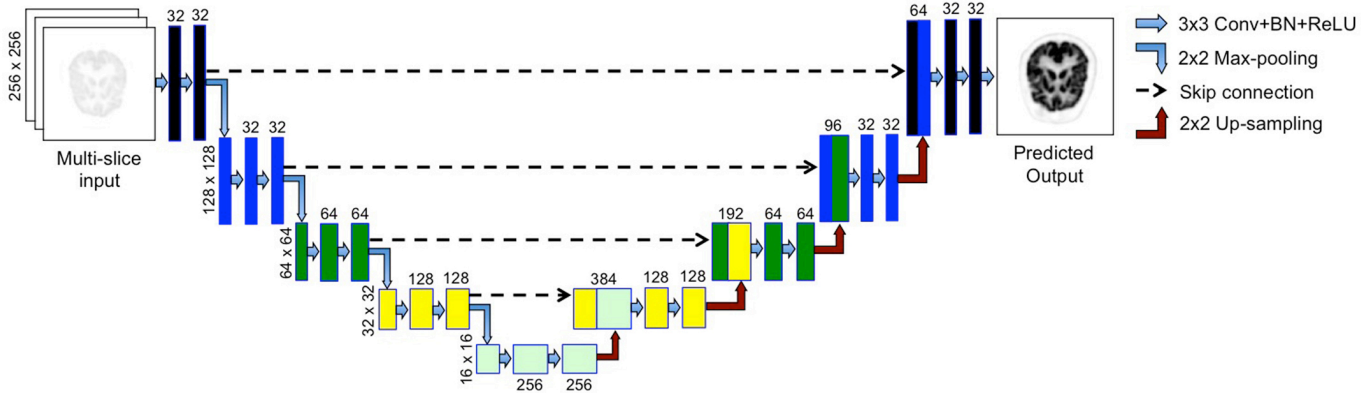

Figure 2. DCNN architecture (Conv: convolution; BN: batch normalization; ReLU: rectified linear unit) with 2M parameters.

(range, 135–900 s). CT-based attenuation/scatter-corrected PET ($PET_{CT-ASC}$) images were reconstructed by a TOF ordered subsets expectation maximization (OSEM) algorithm (4 iterations; 28 subsets; axial FOV, 350 mm; matrix size, 256 × 256 × 89; voxel size, 1.37 × 1.37 × 2.78 mm$^3$; 4.0 mm in-plane Gaussian filter followed by axial 3-slice 1:4:1 filtering) using the offline PET/MR toolbox (REL_1_28, GE Healthcare). Noncorrected PET ($PET_{NC}$) and $PET_{CT-ASC}$ images were reconstructed with other corrections including normalization, dead time, decay, point-spread function, and randoms. $PET_{NC}$ and $PET_{CT-ASC}$ images were utilized as paired input and output for training/testing our proposed DCNN architecture.

$^{18}$F-FDG pancreas PET was used for investigating the adaptability of the developed DCNN architecture for PET abdominal imaging. 20 subjects underwent TOF-PET/MRI (SIGNA, GE Healthcare) for 45 minutes without additional CT. Here, MR-derived pseudo CT were transformed to attenuation maps by the clinical implementation [13]. Pseudo CT-based attenuation/scatter-corrected PET ($PET_{CT-ASC}$) images were reconstructed using a TOF-OSEM algorithm (2 iterations; 28 subsets; matrix size, 256 × 256 × 89; voxel size, 2.34 × 2.34 × 2.78 mm$^3$; 5.0 mm in-plane Gaussian filter followed by axial 3-slice 1:4:1 filtering) using the offline PET/MR toolbox (REL_1_28, GE Healthcare).

### B. DCNN Architecture and Model Training

The proposed DCNN is based on the U-Net architecture [14]. It consists of five encoder-decoder stages with symmetrically concatenated with skip connections (Figure 2). The proposed model was implemented using TensorFlow (version 1.7.0 with CUDA 9.1) and Keras libraries. Training and testing our proposed DCNN were performed on a Ubuntu server (version 16.04 LTS) with a single Tesla P100 (NVIDIA) graphics card. Training parameters and strategies were summarized as follows:
- Preprocessing: Raw values (Bq/ml) were scaled down to kBq/ml, and image slices including the brain were cropped.
- Data augmentation: Random rotation (-10 ~ 10 degree), horizontal flip, and vertical translation (< 50 pixels) were applied.
- Processing in each stage: Convolution (Conv) with 3×3 kernels, batch normalization (BN) [15], and rectified linear unit (ReLU) is sequentially performed twice (Figure 2).
- Processing between stages: Downsampling and upsampling are done by 2×2 max pooling and bilinear interpolation [10], respectively. In order to preserve local information and resolution of the image, skip connections transfer the 2$^{nd}$ convolution layer of the encoder, occurred prior to the BN and ReLU activation, to the decoder after upsampling at the same level of stage [8].
- Loss function and optimizer: Mean squared error (or L2 loss) and RMSprop optimizer [16] were employed with a learning rate initialized by 0.001 which halved automatically if the loss did not decrease in ten epochs.
- Weight initialization: Weights were initialized with truncated Gaussian distributions with zero mean and standard deviation of 0.02.
- All biases were initialized with zero.

A mini-batch of 32 input/output patches was used for training and the loss reached its steady state in 140 epochs (approximately 160 minutes). On the trained model, each prediction of $PET_{DCNN}$ took an average of 0.4 s with the single Tesla P100 graphics card.

### C. Evaluation

For quantitative analysis, the generalized error of our proposed model was quantified by four metrics, including the normalized root mean square error (NRMSE), peak signal to noise ratio (PSNR) and structural similarity index (SSIM). For statistical analysis, joint histogram was used to show the distribution of voxel-based PET uptake correlation between $PET_{CT-ASC}$ and $PET_{DCNN}$ within the SUV (image-derived uptake [MBq/mL] / injection dose [MBq] × patient's weight [g]) range of 0.5–20.0 (g/mL) across test data (ten subjects). Finally, subject-specific differences between CT-ASC and DCNN were illustrated with PET images for selected subjects.

## III. RESULTS

### A. $^{18}F$-FDG Brain PET

The overall performance of $PET_{DCNN}$ is quantitatively comparable to $PET_{CT-ASC}$ (Figure 3). Across the test set of 10 subjects, the NRMSE was 0.099 ± 0.077; the average PSNR was 10.96 ± 4.59; the average SSIM was 0.988 ± 0.007, demonstrating high image similarity between $PET_{DCNN}$ and $PET_{CT-ASC}$. As a reference, the results of the train set show that the performance of DCNN is slightly reduced for the test set.

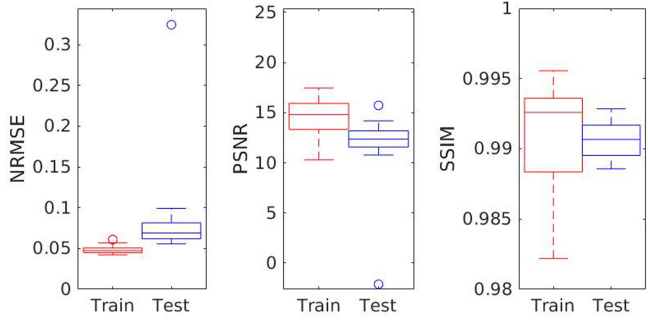

**Figure 3.** Boxplots of the averaged performance (NRMSE; PSNR; SSIM) of $PET_{DCNN}$, compared to reference $PET_{CT-ASC}$.

The joint histogram of voxel-wise PET comparison across the test set shows the voxel-wise similarity of $PET_{DCNN}$ and reference $PET_{CT-ASC}$ with the slope of 1.01 and $R^2$ of 0.98 (Figure 4): $PET_{DCNN}$ achieved higher accuracy for lower uptake voxels with smaller variation but lower accuracy for larger uptake voxels with larger variation.

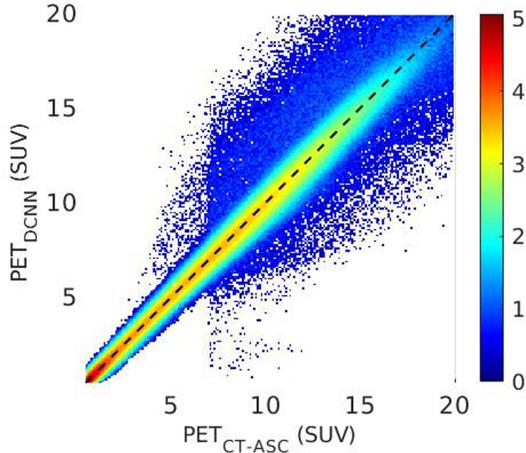

**Figure 4.** Joint histogram of voxel-based PET uptake. The counts were $\log_{10}$-scaled.

Subject-specific similarity and voxel-based difference patterns between CT-ASC and DCNN are illustrated in Figure 5. For subject-1 (top), PET differences are randomly distributed with the mixture of over- and under-estimated patterns. For subject-2 (middle), the $SUV_{max}$ of the tumor with DCNN was underestimated by -13.5%. For subject-3 (bottom), $PET_{DCNN}$ was substantially overestimated but qualitatively similar to $PET_{CT-ASC}$ (SSIM = 0.9670).

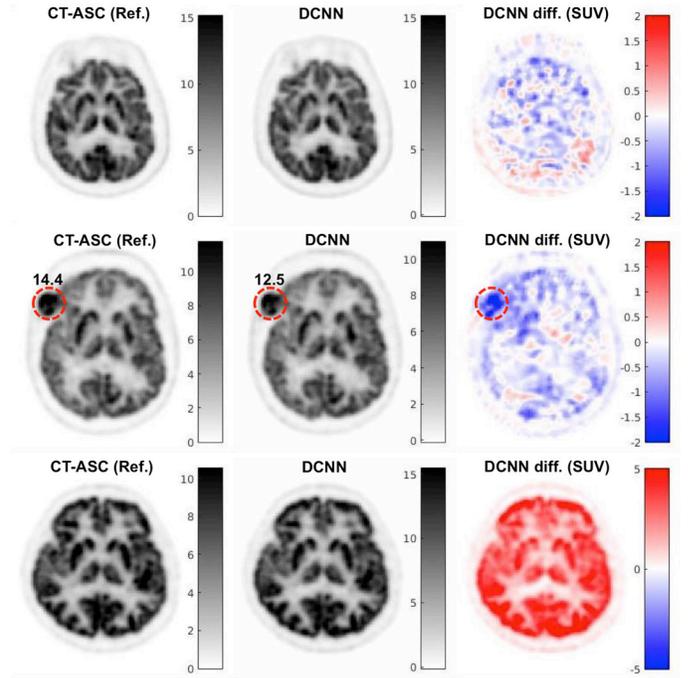

**Figure 5.** PET examples of representative subjects for CT-ASC, DCNN, and their difference: (a) Subject-1 with the longest scan duration (900 s), (b) Subject-2 with a tumor in the head, and (c) Subject-3 with the highest NRMSE.

### B. $^{18}F$-FDG Pancreas PET

In order to investigate the adaptability of our proposed DCNN architecture for another anatomy, the DCNN architecture optimized for PET neuroimaging was applied to the pancreas data. As a preliminary study, all data were used for model training and validation rather than separating training/test data. Since the preliminary result with the pancreas data varied substantially, 20 subjects were grouped into three categories according to the values of PSNR: (a) group-1 (G1, 10 subjects) for PSNR > 10, (b) group-2 (G2, 6 subjects) for 0 < PSNR < 10, (c) group-3 (G3, 4 subjects) for PSNR < 0. Figure 6 shows a large variation between three groups, indicating the limitation of the current DCNN architecture for PET abdominal imaging.

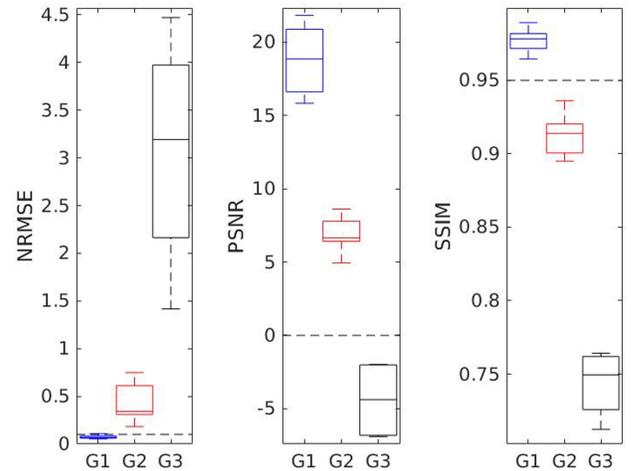

**Figure 6.** Boxplots of the averaged performance (NRMSE; PSNR; SSIM) of $PET_{DCNN}$ for group-1 (G1), group-2 (G2), and group-3 (G3).

Subject-specific similarity and voxel-based difference patterns between CT-ASC and DCNN are illustrated for one subject in group-1 as a successful example (Figure 7).

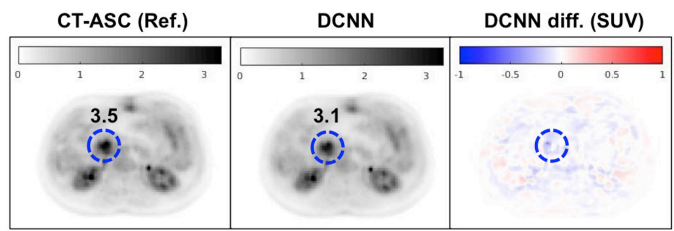

**Figure 7.** PET example of subject-1 for CT-ASC, DCNN, and their difference.

## IV. DISCUSSION

DCNN applications for medical imaging are increasing, inspired by the success of *U-Net* for medical image segmentation [14], which consists of a contracting path and an expansive path to extract features at different resolution. However, DCNN applications for PET are more challenging than those for MR and CT due to the low resolution and noise characteristics of PET. Nevertheless, the success of denoising for low-dose PET [10] demonstrated the capability of DCNN to deal with noisy PET data. In this study, we demonstrated that the proposed DCNN can perform ASC simultaneously for PET neuroimaging in *image* space, which is feasible due to the perceptibility of important structures and their boundaries in both $PET_{NC}$ and $PET_{CT-ASC}$ images.

To our knowledge, this is the first work to investigate the feasibility of joint ASC using DCNN in *image* space. This approach is a one-step process, distinct from conventional methods that rely on generating attenuation maps first that are then applied to iterative scatter simulation in *sinogram* space for quantitative PET image reconstruction.

For the model training, we did not consider the information about a table couch and external head coils that should be always included in attenuation maps derived from CT or MR images for accurate attenuation correction and scatter simulation. Surprisingly, however, the omitted information was not problematic since the attenuation information caused by the external materials could be imbedded in training images themselves.

In particular, our proposed DCNN-based approach has great potential to promote the clinical feasibility for a dedicated brain PET system that needs a practical and robust way for attenuation and scatter correction without requiring an anatomical imaging device such as CT or MRI that can provide attenuation maps. Also, the potential success of DCNN-based ASC for thorax or abdominal PET can be a back-up plan for motion-corrupted attenuation maps.

## V. CONCLUSION

We have demonstrated the feasibility of directly producing PET images corrected for attenuation and scatter using a deep convolutional neural network ($PET_{DCNN}$) from noncorrected PET ($PET_{NC}$) in image space without additional anatomical imaging and time-consuming scatter simulation.

ACKNOWLEDGMENT

The study was supported in part by the National Institutes of Health under Grants R01HL135490 and R01EB026331.